\documentclass[lettersize,journal]{IEEEtran}
\usepackage{amsmath,amsfonts}
\usepackage{algorithmic}
\usepackage{algorithm}
\usepackage{array}
\usepackage[caption=false,font=normalsize,labelfont=sf,textfont=sf]{subfig}
\usepackage{textcomp}
\usepackage{stfloats}
\usepackage{url}
\usepackage{verbatim}
\usepackage{graphicx}
\usepackage{cite}
\usepackage{multirow}
\usepackage{color}
\begin{document}

\title{Semantic Visual Simultaneous Localization and Mapping: A Survey}

\author{

Kaiqi~Chen\textsuperscript{1,2},~Junhao~Xiao\textsuperscript{3},~Jialing~Liu\textsuperscript{1,2},~Qiyi~Tong\textsuperscript{2,4},~Heng~Zhang\textsuperscript{2,4}, \\Ruyu~Liu\textsuperscript{5},~Jianhua~Zhang\textsuperscript{6,*},~Arash Ajoudani\textsuperscript{2}, and Shengyong~Chen\textsuperscript{6}

\thanks{\textsuperscript{*}Corresponding author: Jianhua Zhang, zjh@ieee.org.}

\thanks{\textsuperscript{1}Institute of Computer Vision, College of Computer Science and Technology, Zhejiang University of Technology, Hangzhou 310023, China. chenkaiqi96@outlook.com.} 
\thanks{\textsuperscript{2}Human-Robot Interfaces and Interaction Lab, Istituto Italiano di Tecnologia, Genoa, Italy.}  
\thanks{\textsuperscript{3}National University of Defense Technology, China.}

\thanks{\textsuperscript{4}Università di Genova, Genoa, Italy.}

\thanks{\textsuperscript{5}School of Information Science and Technology, Hangzhou Normal University, Hangzhou 311121, China.}

\thanks{\textsuperscript{6}Institute of Computer Vision, School of Computer Science and Engineering, Tianjin University of Technology, Tianjin 300384, China.}

\thanks{This research was supported by the National Natural Science Foundation of China under Grant 62202137 and Zhejiang Provincial Natural Science Foundation of China under Grant LQ22F030004.}
}

\markboth{Journal of \LaTeX\ Class Files,~Vol.~14, No.~8, August~2021}%
{Shell \MakeLowercase{\textit{et al.}}: A Sample Article Using IEEEtran.cls for IEEE Journals}


\maketitle

\begin{abstract}
Visual Simultaneous Localization and Mapping (vSLAM) has achieved great progress in the computer vision and robotics communities, and has been successfully used in many fields such as autonomous robot navigation and AR/VR. 
However, vSLAM cannot achieve good localization in dynamic and complex environments.
Numerous publications have reported that, by combining with the semantic information with vSLAM, the semantic vSLAM systems have the capability of solving the above problems in recent years.
Nevertheless, there is no comprehensive survey about semantic vSLAM. To fill the gap, this paper first reviews the development of semantic vSLAM, explicitly focusing on its strengths and differences. Secondly, we explore three main issues of semantic vSLAM: the extraction and association of semantic information, the application of semantic information, and the advantages of semantic vSLAM. Then, we collect and analyze the current state-of-the-art SLAM datasets which have been widely used in semantic vSLAM systems. Finally, we discuss future directions that will provide a blueprint for the future development of semantic vSLAM.
\end{abstract}


\section{Introduction}   \label{section1}
\IEEEPARstart{S}{imultaneous} localization and mapping (SLAM) is the foundation for robots to explore the environment autonomously. Humans can quickly locate themselves in unfamiliar and complex environments and reconstruct the environment through spatial perception. This ability can be enhanced with acquired training and plays a crucial role in human cognition and motor control development. 
Similarly, the mobile agents and robots can also estimate information about their motion and the environment, if they equip with different sensors and run the SLAM algorithm. 
SLAM has gradually become synonymous with robotics over the past few decades. In terms of theory and practice, SLAM has now been established a complete set of technical solutions. SLAM technology has been applied widely in many fields, such as Medical Service Robots\cite{yang2020combating}, Autonomous Driving\cite{li2018stereo}, Unmanned Aerial Vehicles (UAVs)\cite{qian2021robust}, Augmented Reality (AR)\cite{liu2021collaborative} and Virtual Reality (VR).

There are various SLAM technology methods, but most are challenging to popularize due to high equipment costs or limited scenarios to be applied. SLAM based on vision sensors, which is also called visual SLAM (vSLAM), has recently become a popular research direction due to its low hardware cost, high accuracy in small scenes, and the ability to obtain rich environmental information. It has to be mentioned that the disadvantages of vSLAM are also very obvious. On the one hand, there are still many challenges in coping with lighting changes, dynamic object movements, and environments lacking textures. On the other hand, the system has a high computing load, and the constructed geometric maps are difficult to apply for path planning and navigation.

The rise of deep learning technologies in recent years has allowed researchers to solve the traditional SLAM problem. Based on deep learning techniques, the researchers extract feature points, descriptors, and semantic information and perform pose estimation.
The integration of semantic information into traditional vSLAM improves the understanding of image features and builds highly accurate semantic maps, which are verified in earlier works\cite{salas2013slam++,li2016semi, ma2017multi, sunderhauf2017meaningful}. 
Compared with traditional vSLAM, semantic vSLAM not only acquires the geometric structure information in the environment but also extracts semantic information about independent objects (e.g., position, orientation, and category). In localization, semantic vSLAM improves the accuracy and robustness of localization with the help of semantic constraints. In mapping, semantic information provides rich object information to build different types of semantic maps, such as pixel-level maps\cite{mccormac2017semanticfusion}, and object-level maps\cite{barsan2018robust,yang2019cubeslam}. 
Therefore, semantic vSLAM can help robots improve the ability to accurately perceive and adapt to unknown complex environments and perform more complex tasks.

Several current survey papers have extensively discussed SLAM algorithms and systems. The early SLAM developments have been well-summarized in\cite{durrant2006simultaneous, bailey2006simultaneous}.
Of course, there are also investigations involving specific domains, such as 
multi-robot collaborative SLAM\cite{saeedi2016multiple}, keyframe-based monocular SLAM\cite{younes2017keyframe}, and SLAM for dynamic environments\cite{saputra2018visual}.
Meanwhile, there is a recent visual SLAM survey in\cite{taketomi2017visual}, where the survey scopes from 2010 to 2016 and lacks a new frontier in semantics. 
Notably, \cite{cadena2016past} provides a detailed review of the tremendous progress which has been made in the SLAM community, and considers the future direction.
While this survey contains a brief discussion of deep learning methods, it does not provide a comprehensive overview of this field, especially the explosion of research over the past seven years.
Therefore, \cite{chen2020survey} introduces the current SLAM combined with deep learning methods. However, it does not summarize semantic data processing and association methods.
\cite{sualeh2019simultaneous} mentions the concept of semantic SLAM but does not emphasize the importance of semantics, especially the contribution made by the rapid development of semantic vSLAM in recent years.
It is worth noting that there have been papers\cite{taketomi2017visual,cadena2016past, chen2020survey,sualeh2019simultaneous} detailing the work related to SLAM, but there are currently no surveys that systematically introduce the development history of semantic vSLAM. Therefore, this survey provides a detailed overview of existing works on semantic vSLAM, focusing on semantic information extraction, semantic applications, SLAM datasets, and comparison of semantic vSLAM with traditional vSLAM. To the best of our knowledge, this is the first survey paper that provides a comprehensive and extensive overview of semantic vSLAM.

\begin{figure*}[htbp]
	\centerline{\includegraphics[width=1.0\textwidth]{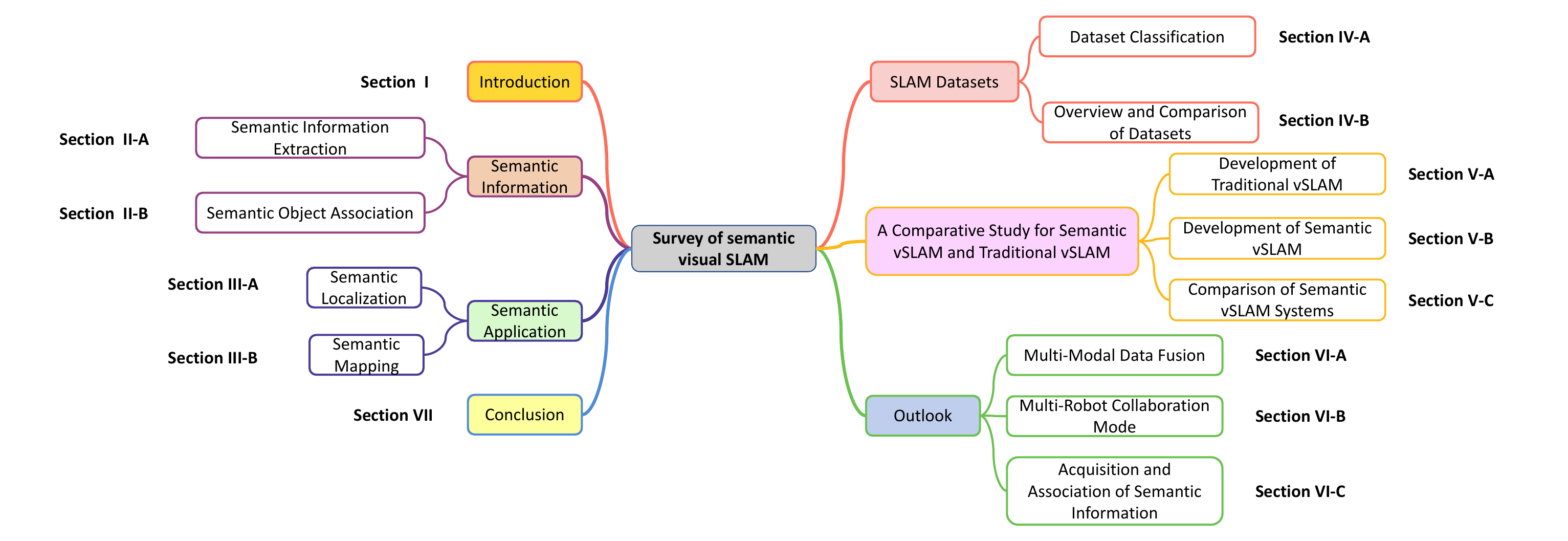}}
	\caption{The schematic diagram of the overall structure of the paper.}
	\label{Fig.1.}
\end{figure*}

The remainder of the paper is structured as shown in Fig.\ref{Fig.1.}. 
Section \ref{section2} introduces the extraction and association methods of semantic information in semantic vSLAM.
Section \ref{section3} introduces the current application of semantic visual SLAM. 
Section \ref{section4} introduces the existing SLAM datasets. 
Section \ref{section5} states the difference between traditional vSLAM and semantic vSLAM, listing and comparing recent semantic vSLAM works. 
Section \ref{section6} introduces the current problems and future development directions of semantic vSLAM technology. 
A summary of semantic vSLAM is given in Section \ref{section7}.

\section{Semantic Information}\label{section2}
In recent years, SLAM has begun to be combined with semantic information, which contains the position, orientation, colour, texture, shape, and specific attributes of objects in the environment. 
Compared with past SLAM methods, Semantic vSLAM not only can acquire geometric structure information in the environment during the mapping process, but also can recognize objects in the environment and acquire semantic information to adapt to complex environments and perform more intelligent tasks. Traditional vSLAM methods are often based on the assumption of a static environment, whereas semantic vSLAM can predict the movable properties of objects in dynamic environments. Similar object knowledge representations in semantic vSLAM can be shared, improving the operation and storage efficiency of SLAM systems by maintaining a shared knowledge base. Moreover, Semantic vSLAM can be applied to intelligent path planning and navigation, such as server robots selecting the optimal path for delivering supplies. 

\begin{figure}[htbp]
	\centerline{\includegraphics[width=0.5\textwidth,height=0.25\textwidth]{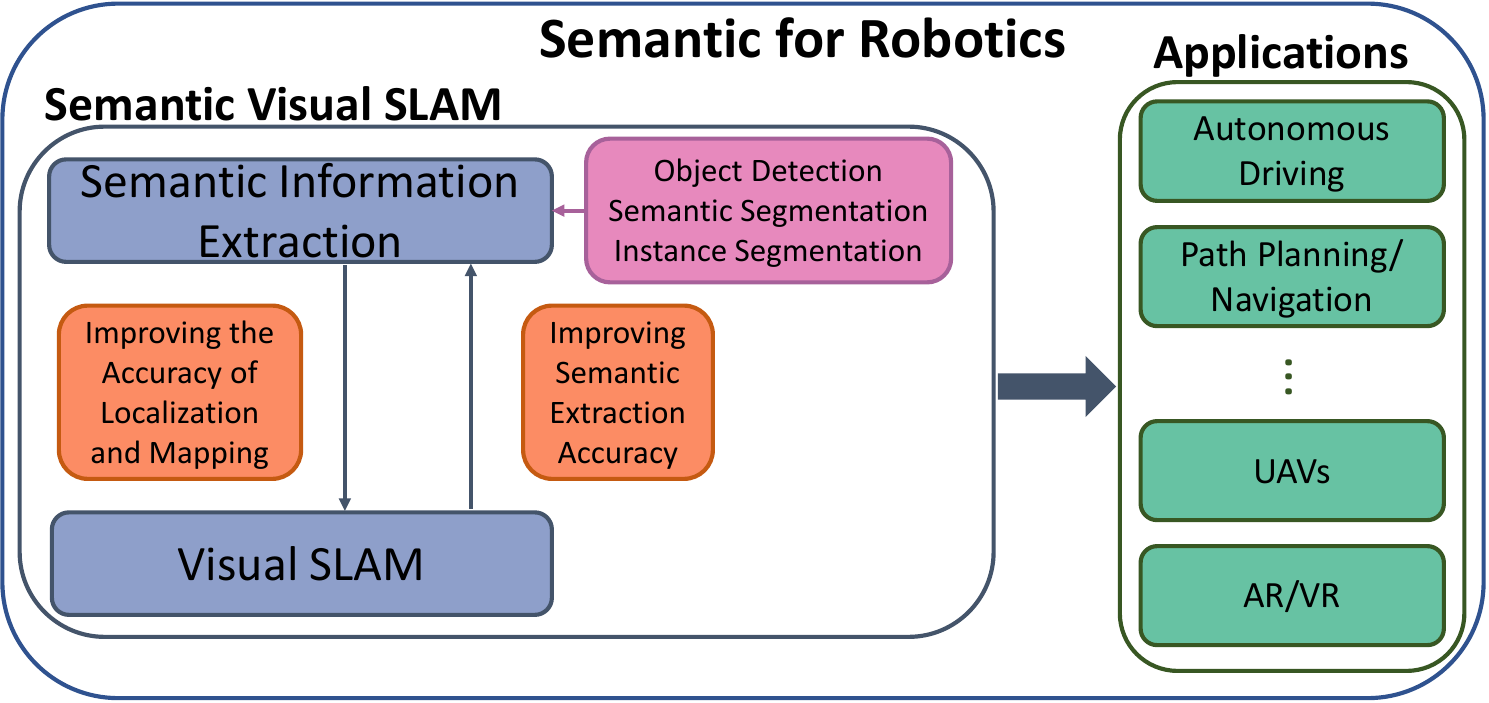}}
	\caption{The overall framework of semantic robotics. Semantic visual SLAM consists of semantic information extraction and visual SLAM modules, which influence each other. Semantic visual SLAM is widely used in autonomous driving, path planning, and navigation.}
	\label{Fig.2.}
\end{figure}

The framework of semantic vSLAM can be roughly divided into the semantic information extraction module and the vSLAM module, as shown in Fig.\ref{Fig.2.}. 
Moreover, the key to semantic vSLAM methods is accurately identifying objects in the environment. 
For the process of semantic extraction, we can regard the process as identifying objects of interest in images and obtaining information about objects. 
The deep learning techniques that have emerged over the years are the most promising methods for semantic extraction. 
The research on semantic object extraction is gradually shifted from traditional machine vision algorithms to the direction of deep neural networks, such as CNN and R-CNN. Their semantic extraction accuracy and real-time performance can meet the requirements of SLAM. There are three commonly used methods for semantic vSLAM extraction of semantic information, namely object detection\cite{liu2016ssd, redmon2016you, redmon2017yolo9000, farhadi2018yolov3, bochkovskiy2020yolov4, girshick2014rich, girshick2015fast,ren2015faster}, semantic segmentation\cite{ronneberger2015u,kendall2015bayesian,badrinarayanan2017segnet,zhao2017pyramid}, and instance segmentation\cite{ he2017mask}. Furthermore, the processing of semantic object association is also crucial. We will describe the extraction and association of semantic information in the following.

\subsection{Semantic Information Extraction} \label{section2.1}

\subsubsection{Object Detection}  \label{section2.1.1}
The object detection module in semantic vSLAM can help the SLAM system to acquire the objects from images. 
By combining it with SLAM, we can construct object-level semantic maps and improve the environment understanding. 
The current object detection methods used in semantic vSLAM are mainly divided into two categories: one-stage method and two-stage method.

Semantic vSLAM usually adopts SDD\cite{liu2016ssd} or YOLO series (YOLOv1\cite{redmon2016you},YOLOv2\cite{redmon2017yolo9000},YOLOv3\cite{farhadi2018yolov3},YOLOv4\cite{bochkovskiy2020yolov4}) as the one-stage object detection methods. 
SSD\cite{liu2016ssd} is the first DNN-based real-time object detector that achieves above 70\% mAP in PASCAL VOC datasets\cite{everingham2010pascal} with 40 FPS in TitanX. SSD is also a one-stage object detection detector that balances speed and accuracy well. Hence, several semantic vSLAM works\cite{zhong2018detect, zhang2019hierarchical, doherty2019multimodal,doherty2020probabilistic} deploy SSD as the detector module. 
\cite{zhang2019hierarchical, doherty2019multimodal,doherty2020probabilistic} utilize SSD to detect static objects and do not consider dynamic objects, but Zhong et al.\cite{zhong2018detect} reliaze that the most semantic vSLAM works perform poorly in dynamic and complex environments. 
To address these challenges, they detect potential moving objects (e.g., people, vehicles, animals) through the SSD detector and cull their regions to eliminate the influence of dynamic objects on pose estimation. Experimental results show that the system facilitates a robot to accomplish tasks reliably and efficiently in unknown and dynamic environments. However, it is inevitable to recognize non-dynamic regions as dynamic regions in violently removing potentially dynamic objects.

The YOLOv1\cite{redmon2016you} detector is earlier than SSD, but its speed and detection performance are comparable to SSD, so it is also applied in semantic vSLAM for object detection\cite{liao2022so}.
Soon Redmon et al. propose YOLOv2\cite{redmon2017yolo9000}, which is a great improvement over the YOLOv1 and SSD in recognition type, detection accuracy, speed, and localization accuracy. Given its good performance enough for semantic vSLAM needs, some works \cite{bavle2018stereo, sucar2018bayesian, jayasuriya2020localising, yang2019cubeslam } try to use the YOLOv2 as the detector for semantic vSLAM. 
Bavle et al.\cite{bavle2018stereo} propose a particle filter localization approach based on semantic information from indoor environments. They fuse semantic information into a prior map, which assists the aerial robot with precise localization. 
In outdoor semantic vSLAM, \cite{sucar2018bayesian,jayasuriya2020localising} use the YOLOv2 to detect outdoor vehicles and street road signs, respectively, obtaining a large number of object measurements for improving localization accuracy.  
Yang et al.\cite{yang2019cubeslam} propose a single-image 3-D cuboid detection method suitable for indoor and outdoor scenes, which acquires 3D semantic objects by the YOLOv2 detector and vanishing point technique.
YOLOv3\cite{farhadi2018yolov3} draws on the residual network structure and multi-scale detection ideas to form a deeper network level, improving mean average precision (mAP) and small object detection.
Therefore, a large number of semantic vSLAM works\cite{ nicholson2018quadricslam, hosseinzadeh2019real, ok2019robust, shan2020orcvio, wu2020eao, qian2021semantic } use this detector to meet the accuracy of object detection and localization in dynamic environments. 
Among these methods, Nicholson et al.\cite{nicholson2018quadricslam} specially design a sensor model for object detector based on YOLOv3. Thus, the challenge of detection of partially visible object is addressed, which substantially improves the accuracy of vSLAM. 
No matter the SSD or YOLO series, they all meet the most crucial requirement of semantic vSLAM, i.e., real-time. Therefore, they are widely used in many sematic vSLAM systems.

In addition to real-time performance, the detection accuracy also influences the performance of semantic vSLAM. Therefore, several works (e.g., \cite{li2018stereo,iqbal2018localization}) adopt the two-stage detectors for object detection, such as R-CNN\cite{girshick2014rich}, Fast R-CNN\cite{girshick2015fast}, Faster R-CNN\cite{ren2015faster}. 
Unlike the one-stage detectors, the two-stage detectors need to obtain the region proposal, classify the results and adjust the candidate bounding box positions. 
Due to the design idea of the two-stage detector, the real-time performance of the two-stage detector is usually slightly worse than that of the one-stage detector, but its detection accuracy is higher than that of the one-stage detector. 
Li et al.\cite{li2018stereo} recognize that Faster R-CNN performs well in detecting small objects, and consequently use the spatial-temporal consistency relationship between semantic information and sparse feature measurements for tracking static and dynamic objects. 
In order to make semantic vSLAM face more scenes, Iqbal et al. \cite{iqbal2018localization} propose the hybrid detector idea based on Fast R-CNN \cite{ren2015faster} and MobileNet \cite{howard2017mobilenets}, where the system can flexibly use different object detectors to cope with object detection in different environments. 
Moreover, Li et al. introduce text objects into the semantic map in \cite{li2020textslam}, where they use the EAST text detector\cite{zhou2017east} and provide directional information for the detected text patches.

Although significant progress has been made in vSLAM and object detection in recent years, the object bounding boxes obtained by object detectors also contain foreground and other object information, which affects object reconstruction and global localization accuracy. Therefore, researchers try to use semantic segmentation or instance segmentation to obtain pixel-level objects.

\subsubsection{Semantic Segmentation} \label{section2.1.2}
Semantic segmentation is the cornerstone technology of image understanding, which can give the exact pixels corresponding to each type of object but cannot distinguish different individuals of the same type. It is pivotal in autonomous driving, UAVs, and wearable device applications. 
The current semantic segmentation methods used in semantic vSLAM are basically based on deep learning methods, such as U-Net\cite{ronneberger2015u}, Bayesian SegNet\cite{kendall2015bayesian}, SegNet\cite{badrinarayanan2017segnet}, PSPNet\cite{zhao2017pyramid}.

U-Net is one of the most commonly used segmentation models, which is simple, efficient, easy to build, and requires only a small datasets for training. 
Therefore, Qin et al. \cite{qin2020avp} classify image pixels into different categories based on the U-Net model, such as lanes, parking lines, speed bumps, and obstacles. Because these classes of objects have clear and stable features, they are used to improve localization accuracy. In addition, parking lines are also used for parking space detection and obstacles are used for path planning. 
SegNet is also often used for semantic segmentation tasks in outdoor environments \cite{murali2017utilizing, yu2018ds,ganti2019network}, the advantages of which are better preservation of image edge information and higher running speed. 
\cite{murali2017utilizing,yu2018ds,ganti2019network} are based on the same framework\cite{mur2017orb}, and each proposes a complete dynamic semantic vSLAM system. The difference is that \cite{murali2017utilizing} and \cite{ganti2019network} cull potential moving objects by default, while \cite{yu2018ds} combines semantic segmentation with the moving consistency checking method to filter out dynamic regions of the scene. Since feature points are removed from these regions, the robustness and accuracy of localization are improved in dynamic scenes. 
Compared with U-Net and SegNet, PSPNet considers the context relationship matching problem, which shows a good segmentation effect even in complex environments. Liu et al. \cite{liu2019global} use the PSPNet to obtain semantic labels of sofas, cupboards, and desks in order to build high-precision semantic scene graphs.

Compared with object detection methods which output coarse detection bounding boxes, semantic segmentation methods can recognize objects at the pixel level, which dramatically helps semantic vSLAM to understand the environment. However, semantic segmentation cannot distinguish object instances from the same category, limiting the application scope.

\subsubsection{Instance Segmentation} \label{section2.1.3}
For detecting dynamic object instances, semantic vSLAM starts using instance segmentation methods, obtaining a pixel-wise semantic segmentation of the images. Instance segmentation is a further refinement of object detection to achieve pixel-level object separation. However, it cannot achieve the same real-time performance as object detection. The current common instance segmentation method used in semantic vSLAM (e.g., \cite{bescos2018dynaslam,runz2018maskfusion,strecke2019fusion,xu2019mid}) is Mask-RCNN \cite{he2017mask}, a powerful image-based instance-level segmentation algorithm that can segment eighty semantic object class labels. 
These works are suitable for dynamic environments because they fuse geometric information with Mask-RCNN to segment dynamic and static objects, obtaining pixel-wise semantic segmentation and instance label information. However, the real-time performance is greatly affected. Compared to traditional vSLAM methods, \cite{bescos2018dynaslam} is more robust to localization in dynamic scenes at the cost of removing dynamic objects. Moreover, \cite{runz2018maskfusion,strecke2019fusion,xu2019mid} not only can track and reconstruct static backgrounds and objects in the scene in real-time, but also have robust tracking accuracy.

Although in current stage, three kinds of semantic extraction methods can meet the basic requirements of semantic vSLAM, there is still much room to be further improved with respect to recognition accuracy and operation speed for semantic extraction methods to be efficiently integrated into the semantic vSLAM systems, as in many complex environment, there are dynamic or occluded objects which will influence the performanc of object detection. To solve these challenges, vSLAM and semantic extraction methods will need to complement each other in the future, helping robots perform more intelligent tasks.

\subsection{Semantic Object Association} \label{section2.2}
When a robot with sensors moves in an unknown environment, it will collect a series of data. In a vSLAM system, the image set $I_{1:T}=\{I_1, \cdots, I_T \}$ contains all images acquired from start time to current time $T$. 
Assuming that the current environment contains object labels $\mathcal{C}=\left \{1, ...,c \right \}$. For example, the common used objects in a semantic vSLAM system are doors, chairs, tables, people, and vehicles. $x_{t}$ represents the camera pose of the image $I_{t}$, which includes the position and orientation. $\mathcal{X}\overset{\triangle }{=} \left \{ x_{t} \right \} ^{T}_{t=1}$ denote the set of camera trajectories at each time. Since the camera pose is incrementally estimated based on the last state rather than directly calculated, the result is susceptible to noise. The camera pose at time $t$ can be expressed as follows:
\begin{equation}
x_{t}=f\left ( x_{t-1},u_{t} \right ) +w_{t},w_{t}\sim \mathcal{N}\left ( 0,R_{t} \right ) , 
\label{equ.1}
\end{equation}
where $u_{t}$ is the motion measurement of vSLAM at time $t$, and $R_{t}$ is the covariance matrix of camera pose noise.

When the robot is at the pose $x_{t}$, it will observe landmark point measurements $y_{t}$ and object measurements $\mathcal{L}_{t}\overset{\triangle }{=} \left \{ L_{tm} \right \} ^{M}_{m=1}$ through the camera. The corresponding camera measurement at time $t$ is as follows:	
\begin{equation}
z_{t}=h\left ( x_{t},y_{t},\mathcal{L}_{t}  \right ) +v_{z},v_{z}\sim \mathcal{N}\left ( 0,Q_{z} \right ) ,
\label{equ.2}
\end{equation}
where $Q_{z}$ is random measurement noise, $\mathcal{Z}\overset{\triangle }{=} \left \{ z_{t} \right \} ^{T}_{t=1}$ is the set of existing sensor measurements at the current time.

The SLAM system creates keyframes $\mathcal{F}_{1:D} \overset{\triangle}{=}\left \{ F_{d} \right \} _{d=1}^{D}$ to reduce repeated compution. 
Assuming that there are some detectable objects in the current scene, each keyframe $F_{d}$ can detect $N_{d}$ object measurements through the object detection method.
Usually, a keyframe $F_{d}$ has multiple object measurements, which can be expressed as a set $\mathcal{L}_{d}\overset{\triangle}{=}\left \{ {L_{di}} \right \} _{i=1}^{N_{d}}$, and 
\begin{equation}
L_{di}=\left \{ L_{di}^{c}, L_{di}^{s},L_{di}^{b}\right \},L_{di}\in \mathcal{L}_{d} ,
\label{equ.3}
\end{equation}
where $L_{di}$ represents the measurement information of the $i$ object in the keyframe $F_{d}$, which usually consists of object category $L_{di}^{c}\in \mathcal{C}$, detection confidence $L_{di}^{s}$, and object detection bounding box $L_{di}^{b}$ in works \cite{bowman2017probabilistic,bavle2018stereo,sucar2018bayesian,zhang2019hierarchical,doherty2019multimodal,doherty2020probabilistic,ok2019robust,qian2021semantic,chen2022accurate}. 
These pieces of information can be obtained from the semantic object extraction methods\cite{liu2016ssd, badrinarayanan2017segnet, he2017mask} in Section \ref{section2.1}.

The number of landmarks in the environment is much smaller than the number of object measurements. The reason is that the same landmarks can be observed in consecutive keyframes, and multiple object measurements are detected in each keyframe. Hence, the concept of object association $S$ is introduced, which specifies that object measurements across keyframes are associated with the same landmark. It can be understood that each object measurement $L_{di}$ is assigned a unique landmark $O_{k}$, i.e., $S_{di}=\left (O_{k},L_{di}\right )$.
There are $d_{t}$ keyframes at $t$ time. The object association at current time $t$ can be expressed as:
\begin{equation}
\mathbb{S}_{t} \overset{\triangle}{=} \left \{ S_{ij} \right \},i\in \left \{  1,...,d_{t}\right \} \enspace and \enspace j\in \left \{ 1,...,N_{d} \right \} ,
\label{equ.4}
\end{equation}

As the robot moves around over time, the number of object measurements accumulates. Thus, object association is a dynamic process that varies with the continuously acquired object measurements. At $t$ time, the object association is estimated in works \cite{bowman2017probabilistic,zhang2019hierarchical,doherty2019multimodal,doherty2020probabilistic,chen2022accurate}: 
\begin{equation}
\mathbb{S}_{t}=arg \max_{\mathbb{S}_{t-1}} p\left \{ \mathbb{S}_{t-1} | X,L,Z \right \} ,X\subseteq \mathcal{X},L\subseteq \mathcal{L} ,Z\subseteq \mathcal{Z} ,
\label{equ.5}
\end{equation}
where $X, L, Z$ are subsets selected from all camera poses $\mathcal{X}$, object measurements $\mathcal{L}$ and sensor measurements $\mathcal{Z}$ by SLAM algorithms, respectively.

Once the object association is implemented, it is necessary to update the camera and object poses with the results of the object and camera optimization \cite{bowman2017probabilistic,zhang2019hierarchical,ok2019robust,doherty2019multimodal,doherty2020probabilistic,chen2022accurate}. 
\begin{equation}
X^{'},L^{'}=arg\max_{X,L} \log_{}{P\left \{ Z|X,L,\mathbb{S}_{t}  \right \} } .
\label{equ.6}
\end{equation}

The above equations describe the basic semantic vSLAM problems. From the above, the role of object association for semantic vSLAM is accurately associating semantic object measurements with object landmarks. The difficulty of object association is correctly associating new object measurements with existing 3D landmarks in the map when there are multiple objects of the same category, similar appearance, and proximity position in the current image. Object association helps the robot to obtain the number of real objects in the environment and integrate them into the semantic map, improving the perception of the environment. Furthermore, it can provide correct optimization constraints for updating the camera poses and object poses. 
Thus, Object association of semantic information is a worthwhile research problem in semantic vSLAM, some researchers consider semantic object association from a probabilistic perspective. 
For example, the probabilistic object association model by Bowman et al.\cite{bowman2017probabilistic} is a milestone work on robust object association in semantic vSLAM. They propose a probabilistic approach to model the object association process and to adopt the EM algorithm to find correspondences between the object measurements and landmarks. Furthermore, they fully consider the ambiguity of object association, which lays the mathematical foundation for the follow-up works of semantic vSLAM.

However, there are many indeterminable factors in real environment, such as a large amount of objects with the same category and close location, that greatly reduce the accuracy of object association. It also has a significant impact on the estimated camera poses and object poses, which leads to larger trajectory errors and lower accuracy of the map. 
To improve association accuracy, researchers try different schemes to solve object association in complex environments. \cite{zhang2019hierarchical} proposes the idea of a hierarchical topic model, which is based on hierarchical Dirichlet object association. They provide a rich basis for object association between object measurements and landmarks by using the information on the position, appearance, and category of object.
Later, the hierarchical object association strategy\cite{chen2022accurate} greatly reduces false associations between objects that are similar in location and appearance, which consists of short-term object association and global object association.
Different from the work of \cite{bowman2017probabilistic}, Doherty et al.  \cite{doherty2019multimodal, doherty2020probabilistic} propose a semantic SLAM approach for probabilistic object association based on approximate max-marginalization, which eliminates the ambiguity of object association variables during inference and retains the standard Gaussian posterior assumptions. Compared with other semantic vSLAM systems, this approach has significant robustness advantages.
In addressing object association between consecutive keyframes, \cite{iqbal2018localization} uses a nonparametric statistical approach, which extracts the regions of object measurements and landmarks in the image and compares the depth similarity of the areas to determine whether they can be associated with each other.

Most probabilistic object association works focus on static objects, but these methods become ineffective in dynamic environments. Given the complexity of dynamic environments, \cite {strecke2019fusion} solves the problems of dynamic object association and occlusion by adopting a probabilistic EM framework. Moreover, they combine depth image and signed distance function methods to improve the accuracy of multi-object tracking. \cite {bescos2021dynaslam} adopts different strategies to associate dynamic and static objects. The association of static object measurement is based on the feature point matching mechanism in \cite{mur2015orb}. For dynamic object association, they consider the two main characteristics of the object at a constant velocity in a short time and feature point matching, then associate the object measurements with landmarks in the map by reprojection method.

Non-probabilistic object association methods are also very popular in object-level semantic vSLAM. For example, in \cite{li2019semantic,hosseinzadeh2019real,jayasuriya2020localising}, the Mahalanobis distance and Hungarian algorithm are used to associate new object measurements with landmarks, but these algorithms consume very large computational resources.
\cite{wu2020eao} builds an integrated object association strategy which integrate parametric and nonparametric statistical tests, as well as IoU-based methods, and takes full advantage of the nature of different statistics.
It is worth mentioning that Wang et al.\cite{wang2021dsp} adopt different object association strategies for different sensor devices. One association strategy is performed in LiDAR mode by comparing the distance between the 3D bounding box and the reconstructed object. Another association strategy is in stereo or monocular camera mode by counting the number of matching feature points between the object measurements and the landmarks.
If multiple object measurements in the current keyframe are associated with the same landmark, the closest object measurement is associated with that landmark, and the others are rejected. If object measurement is not associated with any existing landmark, then it will be initialized as a new landmark. 
\cite{sharma2021compositional} combines geometric and semantic information to propose a hybrid object association method, which enables drift-free tracking without an explicit relocalization module. However, the proposed method needs to improve real-time performance and tracking accuracy in the face of object switches and missed detection.
Unlike the hierarchical object association strategy \cite{chen2022accurate}, the two-step strategy for object association proposed by \cite{qian2021semantic} relies mainly on object category and appearance similarity to match landmarks, which is not applicable in outdoor environments.

Semantic extraction approaches and object association strategies in semantic vSLAM have received extensive attention in recent years. In addition, the current probabilistic object association strategy is one of the best approaches to improve the perception capability of SLAM systems, but its robustness and generality need further improvement. Furthermore, because of the complexity of real-world environments, semantic extraction and object association accuracy are highly susceptible to being affected.

\section{Semantic Application}  \label{section3}
Semantic and SLAM technology are two parts that promote each other. Semantic information combined with localization and mapping can improve localization and scene understanding accuracy. In recent years, semantic vSLAM technology has driven the development of localization and mapping, which has significantly impacted research areas such as autonomous driving, mobile robots, and UAVs.
This section will focus on two aspects of semantic localization and semantic mapping.

\subsection{Semantic Localization}  \label{section3.1}
The purpose of localization is for the robot to obtain its orientation in an unknown environment, that is, to determine its position in the world coordinate system of that environment. Traditional vSLAM is susceptible to environmental factors, resulting in localization failure. Nevertheless, rich semantic information can be extracted in vSLAM, helping vehicles and robots perceive high-level information in the environment. Moreover, there are geometric constraints in the semantic information, which can effectively improve the localization accuracy of the system. Since there are obvious differences between indoor and outdoor environments, resulting in different localization difficulties. Therefore, we state semantic localization from two different environments.

Long-term outdoor visual navigation must face challenges such as long-time operation, cross-weather, and significant light changes. Under these challenges, it is difficult to reliably match features between the image and the map, eventually resulting in poor localization accuracy or even complete failure of the localization algorithm.
To solve these problems, some researchers try a localization algorithm\cite{stenborg2018long} based on semantically segmented images and a semantic point feature map, which solves the problem of long-term visual localization. There are also established medium-term constraints based on the semantic information during tracking in\cite{lianos2018vso}, reducing the drift error of visual odometry. Facing the drastic viewpoint changes, the researchers\cite {gawel2018x} adopt semantic graph descriptor matching for global localization to achieve localization under multiple viewpoints.

The indoor robot localization problem is no less challenging than the outdoor one. vSLAM systems still rely on superficial image information to perceive the environment and lack cognitive-level capabilities. The robustness and reliability of SLAM have not yet reached practicality when entering complex indoor environments with dynamic or significant lighting changes.
In improving the cognitive ability of robots in the environment, \cite{nicholson2018quadricslam} comes up with an object-level semantic vSLAM system, which adopts dual quadrics representation of 3D landmarks for the first time, containing the size, position and orientation of the landmarks. Meanwhile, they derive a factor graph-based SLAM formulation that jointly estimates the dual quadric and camera pose parameters under the assumption of solving object association. 
Similarly, \cite {yang2019cubeslam} is a monocular-based 3D object detection and mapping approach that improves camera pose and reduces monocular drift with the help of semantic object constraints.
EAO-SALM\cite{wu2020eao}, which borrows ideas from \cite{nicholson2018quadricslam} and \cite{yang2019cubeslam}, is a framework for object pose estimation based on iForest. The framework contains an outliers-robust centroid, scale estimation algorithm, and an object pose initialization algorithm, which significantly facilitates the joint pose optimization.
However, \cite{nicholson2018quadricslam, wu2020eao} do not consider dynamic object factors, and ellipsoids or rectangles represent the objects created in the sparse maps without object details. Therefore, the researcher notices that semantic information can help distinguish between static and dynamic objects, improving robot localization accuracy and robustness in dynamic environments. Adopting semantic information to segment moving objects and filter out feature points associated with moving objects \cite{li2017rgb, zhong2018detect,yu2018ds} is one of the frequently implemented approaches, improving the system localization in dynamic environments.

It can be seen that fusing semantic information is the fundamental way to raise robot localization performance. In improving localization, semantic works are often used in the stages of SLAM system initialization, back-end optimization, relocalization, and loop closing. Therefore, efficiently handling and utilizing semantic information is crucial to improving localization accuracy.

\subsection{Semantic Mapping} \label{section3.2}
Mapping is another goal of SLAM, which serves the localization in vSLAM. Usually, we hope the robot saves the map so that the robot does not need to build the map repeatedly in the next work, saving a lot of computational resources. In application, the maps constructed by vSLAM include sparse maps\cite{mur2017orb, campos2021orb}, semi-dense maps\cite{forster2014svo}, and dense maps\cite{whelan2015real,whelan2015elasticfusion}. Compared with sparse maps, dense maps contain many 3D spatial points to describe the map, which is more suitable for localization, navigation, obstacle avoidance, and reconstruction.

However, the traditional vSLAM maps lack high-level environmental semantic information for human-computer interaction, making robots unable to perform complex tasks of intelligent obstacle avoidance, recognition, and interaction. In order to better solve map problems, it is more and more essential to establish an accurate and reliable 3D semantic map. 
Early semantic maps\cite{salas2013slam++} often used a priori object CAD model database to construct 3D semantic maps, which can restore real scenes and save a large amount of space for storing dense point cloud maps. Nevertheless, the CAD models are limited to objects in a pre-defined database. 
In\cite{mccormac2017semanticfusion, tateno2017cnn, mccormac2018fusion++, sharma2021compositional}, the researchers build static dense semantic maps, which integrates dense vSLAM with semantic segmentation labels.
For dynamic environment reconstruction, \cite{runz2018maskfusion, barsan2018robust,zhong2018detect} adopt instance-aware semantic segmentation to classify objects as background, moving objects, or potentially moving objects. However, they failed to achieve the real-time performance of the system.
Considering the real-time problem of semantic vSLAM in mapping, some researchers\cite{nicholson2018quadricslam, qian2021semantic} try to construct sparse semantic maps. Represented by \cite{hosseinzadeh2019real, yang2019cubeslam, zhang2019hierarchical, chen2022accurate, wu2020eao, li2020textslam }, these methods are based on the ORB-SLAM2 framework and combine semantic objects with building sparse 3D semantic object maps in real-time.

It must be noted that semantic maps are more widely used in intelligent scenarios than traditional visual maps. However, it needs to face the challenges of heavy calculation, recognition of different types of objects, and map storage.

\section{SLAM Datasets}   \label{section4}
It is well known that most SLAM systems evaluate their algorithms on multiple public datasets to prove their effectiveness in some aspects due to the expensive equipment and the complexity of device operation. 
The most frequently used SLAM datasets include KITTI\cite{geiger2012we}, TUM RGB-D\cite{sturm2012benchmark}, ICL-NUIM\cite{handa2014benchmark} and EuRoC MAV\cite{burri2016euroc}. 
These datasets are collected from different environments, and suited for different vSLAM algorithms. Therefore, it is extremely important to find appropriate datasets to evaluate a vSLAM. 
Recently, Liu et al.\cite{liu2021simultaneous} collated datasets commonly used in SLAM works in the past decade and provided a comprehensive introduction and analysis of them, which will facilitate the SLAM community to find suitable datasets. However, this survey does not provide a detailed introduction and analysis of the datasets suitable for semantic vSLAM. To fill this gap, we organize datasets suitable for semantic vSLAM, from which we evaluate and compare.

\subsection{Datasets Classification}  \label{section4.1} 
The categorization of datasets is typically based on sensor differences or applicable scenarios to help them understand and utilize existing SLAM datasets. Depending on the sensor, SLAM-related datasets can be divided into LiDAR, vision, and vision-LiDAR fusion datasets. The advantage of vision sensors is that they are inexpensive and ubiquitous vision devices, which can be mobile phones or cameras. Although these devices are not as powerful and accurate as radar devices, they are acknowledged to hold great potential in the SLAM community and are steadily moving forward. Furthermore, semantic-based vSLAM dramatically improves the performance of traditional vSLAM, which is attributed to obtaining rich environmental information in visual images through semantic extraction, helping robots to have a high-level understanding of unknown environments.

To help semantic vSLAM systems choose suitable datasets, we investigate the current open-source datasets and collect thirty-one datasets. Moreover, we provide a dataset selection guide for different vSLAM systems in section \ref{section4.2}. For each dataset, we describe the dataset from eight dimensions, showing as much information about each dataset as possible, as shown in Table \ref{tab1}. We expressly indicate whether the dataset contains semantic annotations. The annotation types are object frame, semantic segmentation, instance segmentation, and whether it is 2D or 3D. Moreover, we also indicate how many object categories are in the datasets because these semantic annotations are very helpful for examining semantic vSLAM. The details of each dimension of the dataset are shown as follows:
$\bullet$ Name: Name of the dataset.

$\bullet$ Year: Year of publication of the dataset.

$\bullet$ Cited: the number of citations in the dataset when investigating the dataset (refer to Google Scholar for the number of citations). 

$\bullet$ Sensors: The camera column contains the typical camera types: color, event, depth, and RGB-D.
The LIDAR/RADAR column contains the standard 2D or 3D LIDAR (with beam numbers ranging from 4, 16, 32, and 64), while some datasets also have radar sensors. 
The IMU column is intended to express whether or not IMU is available in the data.

$\bullet$ Ground Truth: This dimension shows the acquisition of real ground localization information.

$\bullet$ Motion Pattern: The acquired moving platform is given in this dimension, indicating the different motion patterns.

$\bullet$ Environment: This dimension mainly introduces the sequence, length, or scene information of the dataset.

$\bullet$ Annotated information: This dimension mainly introduces whether the dataset contains semantic annotations, which can help semantic vSLAM to verify its performance and provide a large amount of annotation information for training detection models.

\subsection{Overview and Comparison of Datasets}  \label{section4.2}
The results of the dataset collection are shown in Table \ref{tab1}. This paper mainly displays thirty-one datasets that have been open-sourced in recent years, including six classics, highly cited datasets, and twenty-five datasets published in the last five years. Moreover, the paper also details four representative datasets \cite{wang2020tartanair,shi2020we,barnes2020oxford,sheeny2021radiate} in the appendix.

Based on the results of the division in Table \ref{tab1}, This paper provides the following recommended guidelines. 

$\bullet$ Consider various sensor devices. Most of the collected datasets can be used for evaluation for semantic vSLAM, except for MulRan\cite {kim2020mulran}.

$\bullet$ Consider the challenges of the environment (light changes, weak textures, bad weather). Some SLAM systems try to illustrate the robustness of their systems in harsh environments, so researchers can choose datasets\cite{engel2016photometrically, jeong2019complex, wen2020urbanloco, minoda2021viode} for evaluation.

$\bullet$ Consider different scenarios. If researchers need a multi-scene dataset, then they can choose from urban datasets \cite{patil2019h3d, barnes2020oxford}, indoor datasets\cite{saeedi2019characterizing, klenk2021tum}, jungle datasets \cite{wigness2019rugd}.

$\bullet$ Choose datasets with data annotation. If researchers are working on semantic vSLAM, they can choose the evaluation datasets\cite{geiger2012we,wigness2019rugd,kirsanov2019discoman,patil2019h3d,wang2020tartanair,alberti2020idda,pham20203d,cabon2020virtual,keen2021drive,sheeny2021radiate,wang2021cirrus,minoda2021viode}.

$\bullet$ Choose different motion patterns. For different application scenarios, it is necessary to select different motion patterns of capturing equipment, such as robots, cars, UAVs, USVs, handheld devices, and simulation devices.

Recently, the simulation device has been popular with the SLAM community because it does not require consideration of site constraints, and its equipment is less costly and time-consuming. For instance, TartanAir\cite{wang2020tartanair} adopts the Unreal Engine and collects the data using the AirSim plugin developed by Microsoft\cite{shah2018airsim}. Compared with traditional datasets, the datasets collected in simulation contain all kinds of objects, motion diversity, and diverse scenarios.

\begin{table*}[htbp]
	\caption{  Overview and comparison of common and state-of-the-art datasets. }
	\begin{center}
			\renewcommand{\arraystretch}{1.05} 
			\resizebox{2\columnwidth}{!}{ 
	\begin{tabular}{|c|c|c|ccc|c|c|c|c|}
		\hline
		\multirow{2}{*}{Dataset}                                     & \multirow{2}{*}{Year} & \multirow{2}{*}{Cited} & \multicolumn{3}{c|}{Sensors}                                                                                                                & Ground Truth                   & Motion                    & \multirow{2}{*}{Environment}    & \multirow{2}{*}{Annotated information} \\ \cline{4-8}
		&                       &                        & \multicolumn{1}{c|}{Camera}                               & \multicolumn{1}{c|}{LiDAR/Radar}                         & IMU                  & Pose                           & Pattern                   &                                 &                                        \\ \hline
		\multirow{2}{*}{KITTI\cite{geiger2012we}}                    & \multirow{2}{*}{2012} & \multirow{2}{*}{8651}  & \multicolumn{1}{c|}{2*color (stereo)}                     & \multicolumn{1}{c|}{\multirow{2}{*}{1*Velodyne-64}}      & \multirow{2}{*}{N/A} & \multirow{2}{*}{RTK-GPS\&INS}   & \multirow{2}{*}{Car}      & Outdoors                        & 2D \& 3D boxes                          \\
		&                       &                        & \multicolumn{1}{c|}{2*gray (stereo)}                      & \multicolumn{1}{c|}{}                                    &                      &                                &                           & 39.2km                          & 8 categories                           \\ \hline
		\multirow{3}{*}{TUM   RGB-D\cite{sturm2012benchmark}}        & \multirow{3}{*}{2012} & \multirow{3}{*}{2670}  & \multicolumn{1}{c|}{\multirow{3}{*}{1*RGB-D}}             & \multicolumn{1}{c|}{\multirow{3}{*}{N/A}}                & \multirow{3}{*}{N/A} & \multirow{3}{*}{MoCap}         & Handheld/                 & Indoors                         & \multicolumn{1}{l|}{}                  \\
		&                       &                        & \multicolumn{1}{c|}{}                                     & \multicolumn{1}{c|}{}                                    &                      &                                & Wheeled                   & 39  sequences                   & \multicolumn{1}{l|}{}                  \\
		&                       &                        & \multicolumn{1}{c|}{}                                     & \multicolumn{1}{c|}{}                                    &                      &                                & Robot                     &                                 & \multicolumn{1}{l|}{}                  \\ \hline
		ICL-NUIM\cite{handa2014benchmark}                            & 2014                  & 763                    & \multicolumn{1}{c|}{1*RGB-D}                              & \multicolumn{1}{c|}{N/A}                                 & N/A                  & SLAM                           & Handheld                  & Indoors/8 sequences             &                                        \\ \hline
		EuRoC   MAV\cite{burri2016euroc}                             & 2016                  & 989                    & \multicolumn{1}{c|}{2*gray (stereo)}                      & \multicolumn{1}{c|}{N/A}                                 & Y                    & MoCap                          & UAV                       & Indoors/22 sequences            &                                        \\ \hline
		\multirow{2}{*}{TUM   MonoVO\cite{engel2016photometrically}} & \multirow{2}{*}{2016} & \multirow{2}{*}{171}   & \multicolumn{1}{c|}{1*NA-gray}                            & \multicolumn{1}{c|}{\multirow{2}{*}{N/A}}                & \multirow{2}{*}{N/A} & \multirow{2}{*}{Loop Drift}    & \multirow{2}{*}{Handheld} & In-/outdoors                    &                                        \\
		&                       &                        & \multicolumn{1}{c|}{1*WA-gray}                            & \multicolumn{1}{c|}{}                                    &                      &                                &                           & diverse scenes/50 sequences     &                                        \\ \hline
		\multirow{2}{*}{Oxford   RobotCar\cite{maddern20171}}        & \multirow{2}{*}{2017} & \multirow{2}{*}{908}   & \multicolumn{1}{c|}{3*color (stereo)}                     & \multicolumn{1}{c|}{2*Sick-2D}                           & \multirow{2}{*}{N/A} & \multirow{2}{*}{RTK-GPS\&INS}   & \multirow{2}{*}{Car}      & Outdoors/urban                  &                                        \\
		&                       &                        & \multicolumn{1}{c|}{3*fisheye-color}                      & \multicolumn{1}{c|}{1*Sick-4}                            &                      &                                &                           & 100  sequences                  &                                        \\ \hline
		\multirow{2}{*}{Complex   Urban\cite{jeong2019complex}}      & \multirow{2}{*}{2019} & \multirow{2}{*}{101}   & \multicolumn{1}{c|}{\multirow{2}{*}{2*color (stereo)}}    & \multicolumn{1}{c|}{2*Velodyne-16}                       & \multirow{2}{*}{Y}   & RTK-GPS+                       & \multirow{2}{*}{Car}      & Outdoors/urban                  &                                        \\
		&                       &                        & \multicolumn{1}{c|}{}                                     & \multicolumn{1}{c|}{2*Sick-2D}                           &                      & FOG+SLAM                       &                           & diverse scenes                  &                                        \\ \hline
		\multirow{2}{*}{ReFusion\cite{palazzolo2019refusion}}        & \multirow{2}{*}{2019} & \multirow{2}{*}{39}    & \multicolumn{1}{c|}{\multirow{2}{*}{1*RGB-D}}             & \multicolumn{1}{c|}{\multirow{2}{*}{N/A}}                & \multirow{2}{*}{N/A} & MoCap                          & \multirow{2}{*}{Handheld} & Indoors                         &                                        \\
		&                       &                        & \multicolumn{1}{c|}{}                                     & \multicolumn{1}{c|}{}                                    &                      & /laser scanner                 &                           & 26 sequences                    &                                        \\ \hline
		\multirow{2}{*}{RUGD\cite{wigness2019rugd}}                  & \multirow{2}{*}{2019} & \multirow{2}{*}{22}    & \multicolumn{1}{c|}{\multirow{2}{*}{1*RGB}}               & \multicolumn{1}{c|}{\multirow{2}{*}{1*Velodyne-32}}      & \multirow{2}{*}{Y}   & \multirow{2}{*}{GPS+IMU}       & Wheeled                   & Outdoors/jungle                 & 24 categories                          \\
		&                       &                        & \multicolumn{1}{c|}{}                                     & \multicolumn{1}{c|}{}                                    &                      &                                & Robot                     & 18 sequences                    & semantic segmentation                  \\ \hline
		\multirow{2}{*}{DISCOMAN\cite{kirsanov2019discoman}}         & \multirow{2}{*}{2019} & \multirow{2}{*}{10}    & \multicolumn{1}{c|}{1*RGB-D}                              & \multicolumn{1}{c|}{\multirow{2}{*}{1*Simulation}}       & \multirow{2}{*}{Y}   & \multirow{2}{*}{Simulation}    & \multirow{2}{*}{Robot}    & Indoors                         & semantic segmentation                  \\
		&                       &                        & \multicolumn{1}{c|}{1*color (stereo)}                     & \multicolumn{1}{c|}{}                                    &                      &                                &                           & 200 sequences                   &                                        \\ \hline
		\multirow{2}{*}{H3D\cite{patil2019h3d}}                      & \multirow{2}{*}{2019} & \multirow{2}{*}{110}   & \multicolumn{1}{c|}{\multirow{2}{*}{3*color}}             & \multicolumn{1}{c|}{\multirow{2}{*}{1*Velodyne-64}}      & \multirow{2}{*}{Y}   & GNSS+IMU                       & \multirow{2}{*}{Car}      & \multirow{2}{*}{Outdoors/urban} & 3D boxes                               \\
		&                       &                        & \multicolumn{1}{c|}{}                                     & \multicolumn{1}{c|}{}                                    &                      & RTK(GSM)DGPS                   &                           &                                 & 8 categories                           \\ \hline
		\multirow{2}{*}{UZH-FPV\cite{delmerico2019we}}               & \multirow{2}{*}{2019} & \multirow{2}{*}{97}    & \multicolumn{1}{c|}{1*fisheye-RGB (stereo)}               & \multicolumn{1}{c|}{\multirow{2}{*}{N/A}}                & \multirow{2}{*}{Y}   & \multirow{2}{*}{Laser tracker} & \multirow{2}{*}{UAV}      & In-/outdoors                    &                                        \\
		&                       &                        & \multicolumn{1}{c|}{1*event}                              & \multicolumn{1}{c|}{}                                    &                      &                                &                           & 27 sequences                    &                                        \\ \hline
		\multirow{2}{*}{ICL\cite{saeedi2019characterizing}}          & \multirow{2}{*}{2019} & \multirow{2}{*}{10}    & \multicolumn{1}{c|}{RGB-D}                                & \multicolumn{1}{c|}{\multirow{2}{*}{N/A}}                & \multirow{2}{*}{N/A} & \multirow{2}{*}{MoCap}         & MAV                       & Indoors                         & \multirow{2}{*}{}                      \\
		&                       &                        & \multicolumn{1}{c|}{Mono}                                 & \multicolumn{1}{c|}{}                                    &                      &                                & Handheld                  & 16 sequences                    &                                        \\ \hline
		\multirow{4}{*}{EU   Long-Term\cite{yan2020eu}}              & \multirow{4}{*}{2020} & \multirow{4}{*}{38}    & \multicolumn{1}{c|}{\multirow{2}{*}{2*stereo}}            & \multicolumn{1}{c|}{2*Velodyne-32}                       & \multirow{4}{*}{Y}   & \multirow{4}{*}{RTK-GNSS/IMU}  & \multirow{4}{*}{Car}      & Outdoors                        &                                        \\
		&                       &                        & \multicolumn{1}{c|}{}                                     & \multicolumn{1}{c|}{1*4-layer lidar}                     &                      &                                &                           & multi-season                    &                                        \\
		&                       &                        & \multicolumn{1}{c|}{\multirow{2}{*}{2*fisheye-RGB}}       & \multicolumn{1}{c|}{1*1-layer lidar}                     &                      &                                &                           & 63.4km                          &                                        \\
		&                       &                        & \multicolumn{1}{c|}{}                                     & \multicolumn{1}{c|}{1*2D lidar}                          &                      &                                &                           &                                 &                                        \\ \hline
		\multirow{2}{*}{Newer   College\cite{ramezani2020newer}}     & \multirow{2}{*}{2020} & \multirow{2}{*}{36}    & \multicolumn{1}{c|}{\multirow{2}{*}{1*D435i(infrared)}}   & \multicolumn{1}{c|}{\multirow{2}{*}{1*Ouster-64}}        & \multirow{2}{*}{Y}   & 6DOF ICP                       & \multirow{2}{*}{Handheld} & Outdoors/2.2km                  &                                        \\
		&                       &                        & \multicolumn{1}{c|}{}                                     & \multicolumn{1}{c|}{}                                    &                      & Localization                   &                           & 3D reconstruction               &                                        \\ \hline
		\multirow{2}{*}{TartanAir\cite{wang2020tartanair}}           & \multirow{2}{*}{2020} & \multirow{2}{*}{52}    & \multicolumn{1}{c|}{2*color (stereo)}                     & \multicolumn{1}{c|}{\multirow{2}{*}{1*Simulated-32}}     & \multirow{2}{*}{N/A} & \multirow{2}{*}{Simulation}    & \multirow{2}{*}{Random}   & In-/outdoors                    & \multirow{2}{*}{semantic segmentation} \\
		&                       &                        & \multicolumn{1}{c|}{1*depth}                              & \multicolumn{1}{c|}{}                                    &                      &                                &                           & 1037 sequences                  &                                        \\ \hline
		\multirow{2}{*}{CUHK-AHU\cite{chen2020cuhk}}                 & \multirow{2}{*}{2020} & \multirow{2}{*}{2}     & \multicolumn{1}{c|}{\multirow{2}{*}{6*color}}             & \multicolumn{1}{c|}{\multirow{2}{*}{1*VLP-16}}           & \multirow{2}{*}{N/A} & \multirow{2}{*}{GPS+IMU}       & \multirow{2}{*}{Car}      & Outdoors/34 sequences           &                                        \\
		&                       &                        & \multicolumn{1}{c|}{}                                     & \multicolumn{1}{c|}{}                                    &                      &                                &                           & logistics/hill/urban            &                                        \\ \hline
		\multirow{2}{*}{IDDA\cite{alberti2020idda}}                  & \multirow{2}{*}{2020} & \multirow{2}{*}{12}    & \multicolumn{1}{c|}{\multirow{2}{*}{RDB-D}}               & \multicolumn{1}{c|}{\multirow{2}{*}{1*ARS 308-radar}}    & \multirow{2}{*}{N/A} & \multirow{2}{*}{Simulation}    & \multirow{2}{*}{Car}      & Outdoors                        & 24 categories                          \\
		&                       &                        & \multicolumn{1}{c|}{}                                     & \multicolumn{1}{c|}{}                                    &                      &                                &                           & 105 scenarios                   & semantic segmentation                  \\ \hline
		\multirow{2}{*}{OpenLORIS-Scene\cite{shi2020we}}             & \multirow{2}{*}{2020} & \multirow{2}{*}{46}    & \multicolumn{1}{c|}{1*RGB-D}                              & \multicolumn{1}{c|}{1*Hokuyo-2D}                         & \multirow{2}{*}{Y}   & MoCap/                         & Wheeled                   & Indoors/22 sequences            &                                        \\
		&                       &                        & \multicolumn{1}{c|}{2*fisheye-RGB (stereo)}               & \multicolumn{1}{c|}{1*Robosense-16}                      &                      & LiDAR SLAM                     & Robot                     & temporal diversity              &                                        \\ \hline
		\multirow{2}{*}{MulRan\cite{kim2020mulran}}                  & \multirow{2}{*}{2020} & \multirow{2}{*}{51}    & \multicolumn{1}{c|}{\multirow{2}{*}{N/A}}                 & \multicolumn{1}{c|}{1*Ouster-64-3D}                      & \multirow{2}{*}{N/A} & \multirow{2}{*}{FOG+GPS+ICP}   & \multirow{2}{*}{Car}      & Outdoors/urban                  &                                        \\
		&                       &                        & \multicolumn{1}{c|}{}                                     & \multicolumn{1}{c|}{1*Navtech-CIR204-H}                  &                      &                                &                           & 12 sequences                    &                                        \\ \hline
		\multirow{2}{*}{Brno   Urban\cite{ligocki2020brno}}          & \multirow{2}{*}{2020} & \multirow{2}{*}{13}    & \multicolumn{1}{c|}{4*RGB}                                & \multicolumn{1}{c|}{\multirow{2}{*}{2*Velodyne-32}}      & \multirow{2}{*}{N/A} & \multirow{2}{*}{RTK-GNSS/INS}  & \multirow{2}{*}{Car}      & Outdoors/urban                  &                                        \\
		&                       &                        & \multicolumn{1}{c|}{1*thermal camera}                     & \multicolumn{1}{c|}{}                                    &                      &                                &                           & 375.7km/67 sequences            &                                        \\ \hline
		\multirow{2}{*}{A*3D\cite{pham20203d}}                       & \multirow{2}{*}{2020} & \multirow{2}{*}{47}    & \multicolumn{1}{c|}{\multirow{2}{*}{2*color}}             & \multicolumn{1}{c|}{\multirow{2}{*}{1*Velodyne-64}}      & \multirow{2}{*}{N/A} & \multirow{2}{*}{/}             & \multirow{2}{*}{Car}      & Outdoors                        & 3D boxes                               \\
		&                       &                        & \multicolumn{1}{c|}{}                                     & \multicolumn{1}{c|}{}                                    &                      &                                &                           & 55-hours data                   & 7 categories                           \\ \hline
		\multirow{2}{*}{Oxford Radar}                                & \multirow{3}{*}{2020} & \multirow{3}{*}{140}   & \multicolumn{1}{c|}{\multirow{2}{*}{3*color (stereo)}}    & \multicolumn{1}{c|}{1*Radar}                             & \multirow{3}{*}{N/A} & \multirow{3}{*}{GPS\&INS+SLAM}  & \multirow{3}{*}{Car}      & Outdoors/urban                  &                                        \\
		&                       &                        & \multicolumn{1}{c|}{}                                     & \multicolumn{1}{c|}{2*Sick-2D}                           &                      &                                &                           & temporal diversity              &                                        \\
		RobotCar\cite{barnes2020oxford}                              &                       &                        & \multicolumn{1}{c|}{3*fisheye-color}                      & \multicolumn{1}{c|}{2*Velodyne-32}                       &                      &                                &                           & 32 sequences                    &                                        \\ \hline
		\multirow{2}{*}{UrbanLoco\cite{wen2020urbanloco}}            & \multirow{2}{*}{2020} & \multirow{2}{*}{35}    & \multicolumn{1}{c|}{\multirow{2}{*}{1*fisheye-color(HK)}} & \multicolumn{1}{c|}{1*Velodyne-32(HK)}                   & \multirow{2}{*}{Y}   & RTK-GNSS/INS(HK)               & \multirow{2}{*}{Car}      & Outdoors/urban                  &                                        \\
		&                       &                        & \multicolumn{1}{c|}{}                                     & \multicolumn{1}{c|}{1*Robosense-R32(SF)}                 &                      & RTK-GNSS(SF)                   &                           & diverse scenes                  &                                        \\ \hline
		\multirow{2}{*}{Virtual kitti   2\cite{cabon2020virtual}}    & \multirow{2}{*}{2020} & \multirow{2}{*}{96}    & \multicolumn{1}{c|}{2*color (stereo)}                     & \multicolumn{1}{c|}{\multirow{2}{*}{N/A}}                & \multirow{2}{*}{N/A} & \multirow{2}{*}{Simulation}    & \multirow{2}{*}{Car}      & \multirow{2}{*}{Outdoors/urban} & 2D \& 3D boxes/14 categories            \\
		&                       &                        & \multicolumn{1}{c|}{1*depth}                              & \multicolumn{1}{c|}{}                                    &                      &                                &                           &                                 & semantic segmentation                  \\ \hline
		\multirow{2}{*}{TUM-VIE\cite{klenk2021tum}}                  & \multirow{2}{*}{2021} & \multirow{2}{*}{7}     & \multicolumn{1}{c|}{2*uEye}                               & \multicolumn{1}{c|}{\multirow{2}{*}{N/A}}                & \multirow{2}{*}{Y}   & \multirow{2}{*}{MoCap}         & \multirow{2}{*}{Handheld} & Indoors                         &                                        \\
		&                       &                        & \multicolumn{1}{c|}{2*GEN4-CD(event)}                     & \multicolumn{1}{c|}{}                                    &                      &                                &                           & 21 sequences                    &                                        \\ \hline
		\multirow{2}{*}{TUK   Campus\cite{keen2021drive}}            & \multirow{2}{*}{2021} & \multirow{2}{*}{0}     & \multicolumn{1}{c|}{2*stereo}                             & \multicolumn{1}{c|}{1*Ouster-128}                        & \multirow{2}{*}{Y}   & \multirow{2}{*}{GPS}           & \multirow{2}{*}{Car}      & Outdoors                        & 2D boxes/4 categories                  \\
		&                       &                        & \multicolumn{1}{c|}{1*omnidirectional}                    & \multicolumn{1}{c|}{4*2D laser scanners}                 &                      &                                &                           & campus                          & semantic segmentation                  \\ \hline
		\multirow{2}{*}{RADIATE\cite{sheeny2021radiate}}             & \multirow{2}{*}{2021} & \multirow{2}{*}{31}    & \multicolumn{1}{c|}{\multirow{2}{*}{1*color (stereo)}}    & \multicolumn{1}{c|}{1*Radar}                             & \multirow{2}{*}{Y}   & \multirow{2}{*}{GPS+IMU}       & \multirow{2}{*}{Car}      & Outdoors/22 sequences           & 2D boxes                               \\
		&                       &                        & \multicolumn{1}{c|}{}                                     & \multicolumn{1}{c|}{1*Velodyne-32}                       &                      &                                &                           & adverse weather                 & 8 categories                           \\ \hline
		\multirow{2}{*}{Cirrus\cite{wang2021cirrus}}                 & \multirow{2}{*}{2021} & \multirow{2}{*}{4}     & \multicolumn{1}{c|}{\multirow{2}{*}{1*RGB}}               & \multicolumn{1}{c|}{\multirow{2}{*}{2*Luminar Model H2}} & \multirow{2}{*}{Y}   & \multirow{2}{*}{GPS+IMU}       & \multirow{2}{*}{Car}      & Outdoors                        & 3D boxes                               \\
		&                       &                        & \multicolumn{1}{c|}{}                                     & \multicolumn{1}{c|}{}                                    &                      &                                &                           & 12 sequences                    & 8 categories                           \\ \hline
		\multirow{2}{*}{VIODE\cite{minoda2021viode}}                 & \multirow{2}{*}{2021} & \multirow{2}{*}{8}     & \multicolumn{1}{c|}{\multirow{2}{*}{1*color (stereo)}}    & \multicolumn{1}{c|}{\multirow{2}{*}{N/A}}                & \multirow{2}{*}{Y}   & \multirow{2}{*}{Simulation}    & \multirow{2}{*}{UAV}      & In-/outdoors                    & \multirow{2}{*}{instance segmentation} \\
		&                       &                        & \multicolumn{1}{c|}{}                                     & \multicolumn{1}{c|}{}                                    &                      &                                &                           & 12 sequences                    &                                        \\ \hline
		\multirow{2}{*}{USVInland\cite{cheng2021we}}                 & \multirow{2}{*}{2021} & \multirow{2}{*}{1}     & \multicolumn{1}{c|}{\multirow{2}{*}{1*color (stereo)}}    & \multicolumn{1}{c|}{3*Radar}                             & \multirow{2}{*}{Y}   & \multirow{2}{*}{RTK-GNSS/INS}  & \multirow{2}{*}{USV}      & Outdoors/canal                  &                                        \\
		&                       &                        & \multicolumn{1}{c|}{}                                     & \multicolumn{1}{c|}{1*LS C16-16}                         &                      &                                &                           & 26km/33 sequences               &                                        \\ \hline
		\end{tabular}}
\end{center}
\label{tab1}
\end{table*}

\section{A Comparative Study for Semantic vSLAM and Traditional vSLAM}  \label{section5}
The SLAM community has made tremendous progress in the last three decades, and we have witnessed the transition stage of SLAM technology in the industry. Cadena et al.\cite{cadena2016past} survey the development of SLAM over the past three decades and discuss that SLAM is entering a new era, the robust-perception age. Compared with the previous pure geometric vSLAM, the new-stage perceptual SLAM has a more robust performance and a higher level of environmental understanding, which is attributed to the application of image semantic information to SLAM for pose estimation, loop closing, and mapping. Therefore, this section will review the history of vSLAM development and introduce the semantic vSLAM development in recent years.

\subsection{Development of Traditional vSLAM}   \label{section5.1} 
The traditional vSLAM systems estimate the robot poses in unknown environments based on image information and build low-level maps, which use multi-view geometry principles. At present, the traditional vSLAM systems are mainly represented by filtering-based method\cite {davison2007monoslam,mourikis2007multi}, keyframe-based BA method\cite{klein2007parallel,mur2015orb}, and direct tracking method\cite{newcombe2011dtam,engel2014lsd}.
The filter-based vSLAM methods regard the system state at each moment as a Gaussian probability model and help the robot to predict the accurate poses according to the filter. Even with various noises, the filtering always predicts the real motion of the robot. For example, \cite{davison2007monoslam} chooses the extended Kalman filter (EKF). Since the visual SLAM pose estimation problem is not linear, the EKF cannot guarantee the global optimality of pose estimation.
PTAM\cite{klein2007parallel}, as the first keyframe-based BA monocular vSLAM system, lays a foundation for subsequent keyframe-based BA vSLAM works. ORB-SLAM\cite{mur2015orb} is based on PTAM architecture by adding the functions of map initialization and loop closing, and the optimization of keyframe selection and mapping. In localization, its localization error is much smaller than \cite{davison2007monoslam, klein2007parallel}. Soon, the same researching team continuously improves ORB-SLAM and releases open-source vSLAM systems (i.e., ORB-SLAM2 \cite{mur2017orb}, ORB-SLAM3 \cite{campos2021orb}). The localization accuracy of these systems is much higher than \cite {whelan2015elasticfusion, whelan2015real, kerl2013dense}. 
Direct tracking methods (i.e., DTAM\cite{newcombe2011dtam}, LSD-SLAM\cite{engel2014lsd}) do not rely on the extraction and matching of feature points, but solve the camera motion by constructing the photometric error from the pixel gray values between the front and back frames. 
In the case of missing features and blurred images, these methods have better robustness than the previous two methods. However, the direct tracking methods are more sensitive to illumination changes and dynamic interference, so the positioning accuracy is generally inferior to \cite{ mur2015orb, mur2017orb}.

\begin{figure}[htbp]
	\centerline{\includegraphics[width=0.5\textwidth]{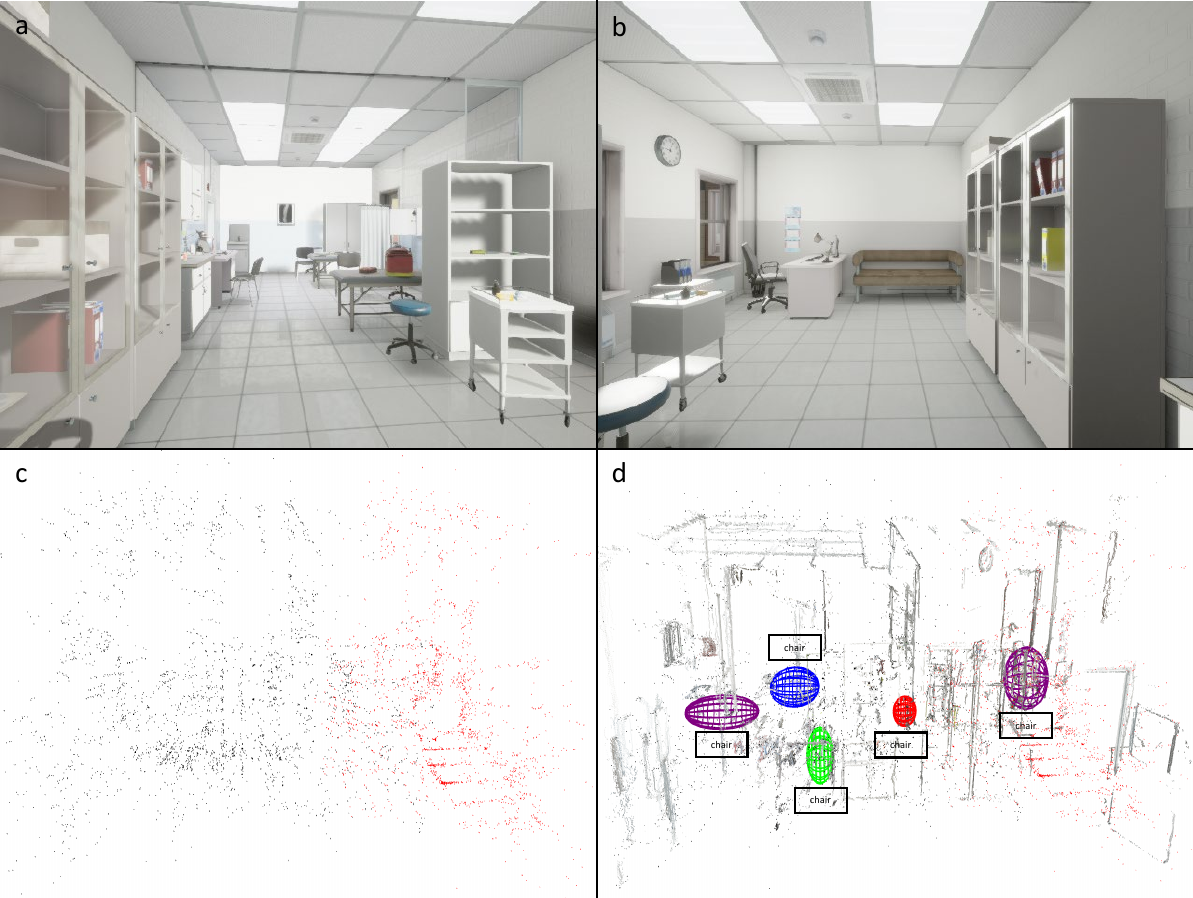}}
	\caption{(a)(b) Images of scenes from different perspectives. (c) 3D map based on point cloud representation of traditional vSLAM. (d)Environment reconstruction with semantic information.}
	\label{Fig.3.}
\end{figure}

As introduced in Section \ref{section3.2}, traditional vSLAM represents the surrounding environment through point clouds, such as sparse maps, semi-dense maps, and dense maps. Since the point clouds in these maps do not correspond to objects in the environment, they are meaningless to the robot, as shown in Fig.\ref{Fig.3.}c. Therefore, researchers try to use geometric and a priori perception information to condense the features of 3D point clouds and understand them, which helps robots perceive high-level environmental details. Coinciding with the rise of semantic concepts, vSLAM systems combine with semantic information solutions greatly improve the ability of robots to perceive the unexplored environment.
In Fig.\ref{Fig.3.}d, semantic information gives semantic labels to point clouds, helping build a semantic map of 3D landmark information. After years of development and verification, semantic information has improved the robustness of vSLAM to the environment and achieved more accurate loop closure.

\subsection{Development of Semantic vSLAM}   \label{section5.2}
The early works of semantic vSLAM can be traced back to the early investigation by Salas-Moreno et al. \cite{salas2013slam++}. They realize that traditional vSLAM systems operate at the low-level primitives which need to be processed harshly. Furthermore, building Maps by traditional methods is only noisy point clouds that do not appear like the maps shown in human vision. It is worth noting that the system is limited to a pre-established object database and has strict requirements on the location of the detected objects. However, it provides the necessary foundation for the subsequent development of semantic vSLAM.
In recent years, the feature point-based vSLAM system has shown outstanding accuracy and robustness in localization, so researchers have tried to build a semantic vSLAM system based on the ORB-SLAM2 algorithm framework. 
For example, researchers introduce 3-D rectangular into the map to construct a lightweight semantic map in \cite{yang2019cubeslam,zhang2019hierarchical,chen2022accurate}.
Other researchers\cite{rubino20173d, nicholson2018quadricslam, ok2019robust} adopt semantic 3-D ellipsoids to build semantic maps because of their ability to compactly represent the size, location, and orientation of landmarks.
Liao et al. incorporate three spatial structure constraints based on \cite{nicholson2018quadricslam}, and propose a monocular object SLAM algorithm for indoor environments with fully coupled three spatial structure constraints\cite{liao2022so}. 
Soon after, EAO-SLAM\cite{wu2020eao} integrates the methods of \cite{nicholson2018quadricslam, yang2019cubeslam} and improves the object pose estimation based on the iForest method, making it possible to estimate the position, pose and scale of landmarks more accurately.

However, previous works on semantic vSLAM often assume that the scenes are static, which limits its application scenarios. These methods do not obtain robustness in localization and mapping when facing dynamic environments. 
To solve the challenge, some researchers propose dynamic outlier detection strategies to remove dynamic objects. 
For example, the motion consistency check algorithm based on semantic segmentation \cite{yu2018ds}, the multiview geometry algorithm based on semantic segmentation \cite{bescos2018dynaslam}, and the moving object detector \cite{zhong2018detect}.
\cite{brasch2018semantic} use prior semantic information to build an efficient online probabilistic model for monitoring dynamic outliers.
In \cite{bescos2019empty,bescos2020empty,besic2022dynamic}, they turn dynamic object regions into realistic static images, improving vision-based localization and mapping tasks in dynamic environments. 
Moreover, other researchers focus on object motion tracking and pose estimation in the literature\cite{huang2019clusterslam,henein2020dynamic,zhang2020robust}. \cite{zhang2020vdo} integrates their previous works\cite{henein2020dynamic,zhang2020robust} and proposes a novel feature-based dynamic SLAM system that leverages semantic information to localize the robot, build the environment structure, and track motions of rigid objects.
They rely on a denser object feature to ensure more robust tracking than \cite{henein2020dynamic}, and their object tracking accuracy is much better than \cite{zhang2020robust}, due to their method can track occluded objects.

In summary, the development of semantic vSLAM has received much attention in recent years, but many solutions are limited to specific scenarios and face many challenges for practical applications. When facing with processing a large number of semantic object measurements in a short period, effective filtering and association of semantic information are still worth further research.

\begin{table*}[htb]
	\caption{  Comparison of the properties of Semantic visual SLAM systems. }
		\begin{center}
	\renewcommand{\arraystretch}{1.2} 
	\begin{tabular}{|c|c|c|c|c|c|c|c|c|}
		\hline
		Method                                     & Year & Input        & Full Shape & \begin{tabular}[c]{@{}c@{}}Detailed\\      Shape\end{tabular} & Large Scene & Dynamic Scene & FPS   & online     \\ \hline
		Co-Fusion\cite{runz2017co}                 & 2017 & RGB-D        & \checkmark           & \checkmark                                                    &             & \checkmark    & /     & \checkmark \\ \hline
		MaskFusion\cite{runz2018maskfusion}        & 2018 & RGB-D        & \checkmark           & \checkmark                                                    &             &               & 30    & \checkmark \\ \hline
		Fusion++\cite{mccormac2018fusion++}        & 2018 & RGB-D        & \checkmark           & \checkmark                                                    &             &               & 4-8   & \checkmark \\ \hline
		DS-SLAM\cite{yu2018ds}                     & 2018 & RGB-D        & \checkmark            & \checkmark                                                    &   & \checkmark    & /     & \checkmark \\ \hline
		Detect-SLAM\cite{zhong2018detect}          & 2018 & RGB-D        & \checkmark           & \checkmark                                                    &   & \checkmark    & /     & \checkmark \\ \hline
		DynaSLAM\cite{bescos2018dynaslam}          & 2018 & Multiple     & \checkmark           & \checkmark                                                              & \checkmark  & \checkmark    & /     & \checkmark \\ \hline
		DynSLAM\cite{barsan2018robust}             & 2018 & Stereo       & \checkmark & \checkmark                                                    & \checkmark  & \checkmark    & 2     &            \\ \hline
		Li et al.\cite{li2018stereo}               & 2018 & Stereo       & \checkmark & \checkmark                                                    & \checkmark  & \checkmark    & 5.8   & \checkmark \\ \hline
		EM-Fusion\cite{strecke2019fusion}          & 2019 & RGB-D        & \checkmark & \checkmark                                                    &             & \checkmark    & /     & \checkmark \\ \hline
		MID-Fusion\cite{xu2019mid}                 & 2019 & RGB-D        & \checkmark & \checkmark                                                    &             & \checkmark              & 2-3   & \checkmark \\ \hline
		QuadricSLAM\cite{nicholson2018quadricslam} & 2019 & RGB          & \checkmark &                                                               & \checkmark  &               & /     & \checkmark \\ \hline
		CubeSLAM\cite{yang2019cubeslam}            & 2019 & RGB          & \checkmark &                                                               & \checkmark  &               & 10-30 & \checkmark \\ \hline
		Liu et   al.\cite{liu2019global}           & 2019 & RGB-D        & \checkmark            & \checkmark                                                    &             & \checkmark    & /     & \checkmark \\ \hline
		HDP-SLAM\cite{zhang2019hierarchical}       & 2019 & RGB, RGB-D    & \checkmark &                                                               & \checkmark  &               & /     & \checkmark \\ \hline
		Deep-SLAM++\cite{hu2019deep}               & 2019 & RGB-D        & \checkmark            & \checkmark                                                    &             &               & /     & \checkmark \\ \hline
		ClusterSLAM\cite{huang2019clusterslam}     & 2019 & Stereo       &            &                                                               & \checkmark  & \checkmark    & 7.1   & \checkmark \\ \hline
		DXSLAM\cite{li2020dxslam}                  & 2020 & RGB-D        &            &                                                               & \checkmark  &               & 21.6  & \checkmark \\ \hline
		AVP-SLAM\cite{qin2020avp}                  & 2020 & RGB+IMU      & \checkmark           &                                                     & \checkmark  &               & 15    & \checkmark \\ \hline
		NodeSLAM\cite{sucar2020neural}             & 2020 & RGB-D        & \checkmark & \checkmark                                                    &   &               & /     & \checkmark \\ \hline
		EAO-SLAM\cite{wu2020eao}                   & 2020 & RGB          & \checkmark & \checkmark                                                    &   &               & /     &            \\ \hline
		TextSLAM\cite{li2020textslam}              & 2020 & RGB          & \checkmark & \checkmark                                                    & \checkmark  &               & /     & \checkmark \\ \hline
		ClusterVO\cite{huang2020clustervo}         & 2020 & Stereo       & \checkmark & \checkmark                                                    & \checkmark  & \checkmark              & 8     & \checkmark \\ \hline
		VDO-SLAM\cite{zhang2020vdo}                & 2020 & Stereo, RGB-D & \checkmark          & \checkmark                                                    & \checkmark  & \checkmark    & 5-8   & \checkmark \\ \hline
		Empty   Cities\cite{bescos2020empty}       & 2021 & RGB          & \checkmark            &                                                               & \checkmark  & \checkmark    & /     & \checkmark \\ \hline
		DSP-SLAM\cite{wang2021dsp}                 & 2021 & Multiple     & \checkmark & \checkmark                                                    & \checkmark  &               & 10-20 & \checkmark \\ \hline
		DynaSLAM   II\cite{bescos2021dynaslam}     & 2021 & Stereo, RGB-D & \checkmark & \checkmark                                                    & \checkmark  & \checkmark    & 10-12 & \checkmark \\ \hline
		Qian et   al.\cite{qian2021semantic}       & 2021 & RGB-D        & \checkmark &                                                               &             &               & 30    & \checkmark \\ \hline
		Sharma et   al.\cite{sharma2021compositional}  & 2021 & RGB-D        & \checkmark            & \checkmark                                                    &             &               & /     & \checkmark \\ \hline
		SO-SLAM\cite{liao2022so}                   & 2022 & RGB          & \checkmark &                                                               & \checkmark  &               & /     & \checkmark \\ \hline
		Chen et   al.\cite{chen2022accurate}       & 2022 & RGB, RGB-D    & \checkmark &                                                               & \checkmark  &               & /     & \checkmark \\ \hline

	\end{tabular}
\label{tab2}
\end{center}
\end{table*}

\subsection{Comparison of Semantic vSLAM Systems}  \label{section5.3}
To compare semantic vSLAM more graphically, we collect thirty semantic vSLAM systems from 2017 to 2022, as shown in Table \ref{tab2}. For each semantic vSLAM system, we describe it in nine basic dimensions, which can reveal the advantages and disadvantages of the system. The details of each dimension are shown below. Method: name of the semantic vSLAM system. Year: the year in which the method was published. Input: type of sensor used by semantic vSLAM. Full Shape: can the object be reconstructed completely? Detailed Shape: Is it possible to know the detailed Shape of the reconstructed object? Large Scene: Can it be used for large scenes? Dynamic Scene: Can it be used for dynamic scenes? FPS: Semantic vSLAM run rate(/ means the run rate is unknown). Online: Can the system run online?

As shown in the table \ref{tab2}, different semantic vSLAM systems have their characteristics. For example, The advantage of \cite{bescos2018dynaslam, wang2021dsp} is that it is suitable for various types of sensors, meeting real-time needs. \cite{ li2018stereo,barsan2018robust, xu2019mid,zhang2020vdo,bescos2021dynaslam} show advantages in reconstructing 3D objects, making up for the shortcomings of \cite{nicholson2018quadricslam,yang2019cubeslam,huang2019clusterslam,li2020dxslam}. However, reconstructing objects consumes many computing resources, making it potentially inferior to other semantic vSLAM systems in real-time performance. \cite{huang2019clusterslam, huang2020clustervo,zhang2020vdo, bescos2021dynaslam} are applicable for dynamic outdoor scenes, and the robustness of these systems is much better than indoor or static vSLAM. From the semantic vSLAM works between 2017 and 2022, the input data types of semantic vSLAM systems are becoming increasingly multi-modal, and these works have an increasing emphasis on object reconstruction. Furthermore, the application scenes of these works are gradually developing toward large-scale and dynamic environments. The real-time performance of semantic vSLAM can meet the requirements of current applications without reconstructing the appearance of objects.

Looking at the development of SLAM in recent years, semantic vSLAM has been considered the best approach to improve the perception capability of vSLAM systems. Traditional vSLAM systems mainly use low-level geometric features for matching and localizing, such as corner points, lines, and surface features. With the introduction of semantic information, the semantic vSLAM system can perceive advanced information about the environment, such as identifying pedestrians and detecting vehicles, which greatly enriches map information and improves the localization accuracy. Currently, semantic information can be used in all stages of the traditional SLAM algorithm framework, including initialization, back-end optimization, relocalization, and loop closing.
However, the contradiction between the current limited computational resources and the increasing demand for computational resources of algorithms greatly hinders the development of semantic vSLAM. For instance, in semantic information extraction, the systems need to obtain real-time semantic information and require timely filtering and associating of semantic information. In addition, it should be noted that semantic vSLAM is still in the development stage, and many hidden problems need to be solved. For example, wrong object association will make the systems more vulnerable in object-level SLAM.

\section{Outlook}   \label{section6}

\subsection{Multi-Modal Data Fusion}    \label{section6.1}
Some semantic SLAM works use multi-modal sensors (e.g., RGB cameras, Depth cameras, LiDAR) for pose estimation and mapping in unknown environments. 
Multi-modal semantic SLAM systems can be more robust and accurate in complex and dynamic environments. Because these systems incorporate multi-modal semantic information, reducing the ambiguity of object associations. Moreover, these systems more accurately recognize dynamic objects to reduce localization drift caused by dynamic objects. Of course, the processed multi-modal environmental information can be used to construct dense semantic maps. 
However, in complex and highly dynamic environments, the semantic information acquired by these sensors alone is no longer sufficient for real needs. Therefore, future semantic SLAM works can try to fuse more sensors (e.g., Millimeter-wave radar, Infrared cameras, and Event cameras) and prior semantic maps (e.g., 2.5D maps). While multi-modal approaches can obtain richer semantic object information and help improve the ambiguity of object associations, they also bring challenges, such as calibration and synchronization of multiple sensors, real-time fusion, and the association of multi-modal semantic information.

\subsection{Multi-Robot Collaboration Mode}    \label{section6.2}
Multi-robot systems are one of the most important research directions in robotics. In the multi-robot cooperative SLAM system, mutual communication and coordination among robots can effectively utilize spatially distributed information resources and improve problem-solving efficiency. Moreover, the damage of a single robot in the system will not affect the operation of other robots, which have better fault tolerance and anti-interference than single-robot systems. In traditional SLAM research, there are two collaboration methods for multi-robot systems. One is that each client robot builds a local map individually, and the server receives and fuses all the local maps to build a globally consistent map. The other is a decentralized architecture. 
The premise of multi-robot collaboration is how to efficiently and accurately perform multi-robot global localization, but the appearance-based localization methods are difficult to achieve accurate localization under significant viewpoint differences and light changes. Recently, the fusion of semantic information (e.g., text information) helps the multi-robot system to be more robust, which is attributed to the appearance and context-based semantic localization methods that can perform global localization stably and accurately. In addition, multi-robots bring multi-view semantic information for semantic vSLAM. For example, in object association, observing objects from a multi-view increases the number of observations of the same object, which can effectively avoid the ambiguity problem of object association. But it also increases the computational cost simultaneously.

\subsection{Acquisition and Association of Semantic Information}    \label{section6.3}
The acquisition and association of semantic information is still a problem worthy of research in semantic vSLAM systems. The current semantic information acquisition method is based on the deep learning model, and the generalization and accuracy of the model determine the accuracy of the semantic information. For example, when an object is occluded, it is easily ignored by object detection methods. As the number of object measurements accumulates, it becomes more difficult to associate object measurements to landmarks correctly. Current object association methods are often based on semantic information such as distance, orientation, and appearance. However, it is impossible to accurately associate objects by adopting the conventional methods when objects of the same category, close to each other, obscured objects or dynamic objects appear in the environment. Therefore, we need more in-depth research to mine potential semantic information constraints which can improve object association and global localization.

\section{Conclusion}   \label{section7}
This survey summarizes the recent developments in semantic information for robot vision perception, involving semantic information extraction, object association, localization, and mapping. To give the reader an overview of the current state of the field, we summarize some representative works in the survey. We introduce three types of deep learning model-based methods for obtaining semantic information: object detection, semantic segmentation, and instance segmentation. We also introduce the problem of semantic information association. We summarize the application of semantic information in vSLAM. Moreover, we collect and evaluate thirty open-source SLAM datasets. Finally, we present the differences between traditional and semantic vSLAM, listing thirty semantic vSLAM systems. Most of the references in the survey are from the last five years, and we also provide some views on the future development of semantic vSLAM.



\bibliographystyle{IEEEtran}
\bibliography{bare_jrnl_new_sample4}


\end{document}